\documentclass[11pt,a4paper]{article}
\usepackage[hyperref]{emnlp2020}
\usepackage{times}
\usepackage{latexsym}

\usepackage{amsmath}

\usepackage{booktabs}       %
\usepackage[utf8]{inputenc} %
\usepackage{url}
\usepackage{graphicx}
\usepackage{caption}
\usepackage{multirow}
\usepackage{tablefootnote}
\usepackage{colortbl,multirow,hhline}

\usepackage[ruled,vlined]{algorithm2e}

\usepackage{xcolor}
\usepackage{pifont}
\usepackage{fancybox}
\usepackage{paralist, tabularx}

\usepackage{hhline}

\usepackage{bold-extra}

\usepackage{microtype}

\aclfinalcopy %

\newcommand*\colourcheck[1]{%
  \expandafter\newcommand\csname #1check\endcsname{\textcolor{#1}{\ding{52}}}%
}
\colourcheck{green}

\newcommand*\colourX[1]{%
  \expandafter\newcommand\csname #1X\endcsname{\textcolor{#1}{\ding{55}}}%
}
\colourX{red}
\newcommand{\lxmert}{\mbox{\sc{Lxmert}}}
\newcommand{\xlxmert}{\mbox{\sc{X-Lxmert}}}
\newcommand{\uniter}{\mbox{\sc{Uniter}}}
\newcommand{\xuniter}{\mbox{\sc{X-Uniter}}}
\newcommand{\vilbert}{\mbox{\sc{Vilbert}}}
\newcommand{\hummus}{\mbox{\sc{Hummus}}}
\newcommand{\coco}{\mbox{\sc{Coco}}}
\newcommand{\nlvr}{NLVR\textsuperscript{2}}

\title{X-LXMERT: Paint, Caption and Answer Questions \\with Multi-Modal Transformers}

\author{
    Jaemin Cho$^{1,2}$\thanks{\ \ This work was done as part of the Pre-Doctoral Young Investigator residency program at the Allen Institute for AI.}
    \quad
    Jiasen Lu$^1$
    \quad
    Dustin Schwenk$^1$
    \quad
    Hannaneh Hajishirzi$^{1,3}$
    \quad
    Aniruddha Kembhavi$^{1,3}$ \\
    Allen Institute for AI$^1$
    \quad
    UNC Chapel Hill$^2$
    \quad
    University of Washington$^3$
    \\
    { \tt jmincho@cs.unc.edu \quad \{jiasenl,dustins,hannah,anik\}@allenai.org } \\
    \\
    \emph{\small{Code and Demo:}} \url{https://prior.allenai.org/projects/x-lxmert}
}

\date{}

\begin{document}
\maketitle

\begin{abstract}
Mirroring the success of masked language models, vision-and-language counterparts like \vilbert, \lxmert\ and \uniter\ have achieved state of the art performance on a variety of multimodal discriminative tasks like visual question answering and visual grounding. Recent work has also successfully adapted such models towards the generative task of image captioning. This begs the question: \emph{Can these models go the other way and generate images from pieces of text?} Our analysis of a popular representative from this model family -- \lxmert\ -- finds that it is unable to generate rich and semantically meaningful imagery with its current training setup. We introduce \xlxmert, an extension to \lxmert\ with training refinements including: discretizing visual representations, using uniform masking with a large range of masking ratios and aligning the right pre-training datasets to the right objectives which enables it to paint. \xlxmert's image generation capabilities rival state of the art generative models while its question answering and captioning abilities remains comparable to \lxmert. Finally, we demonstrate the generality of these training refinements by adding image generation capabilities into \uniter\ to produce \xuniter.
\end{abstract}
\section{Introduction}

The past year has seen a spate of BERT-style~\cite{Devlin2019} transformer-based architectures  \cite{Lu2019,Chen2019,Li2019} proposed for vision-and-language tasks. These models are typically pre-trained on large image captioning corpora, extending ideas from masked language modeling to mask both the image and text modalities and produce state of the art results on a variety of vision and language tasks including visual question answering, visual grounding and image retrieval. 
These impressive results as well as recent probing mechanisms~\cite{Ilharco2020} suggest that these models are able to capture a variety of semantics in images including objects, attributes and their relationships and ground these in natural language.

While these models have been extensively evaluated over several discriminative tasks, relatively little attention has been paid to their generative capabilities. Bidirectional transformer models like BERT which exploit context preceding and following the current token are not explicitly designed for generation. Recent work for language-only transformers~\cite{Wang2019, Dong2019, Liao2020} adapt these models towards this capability using sampling procedures.  Such techniques have also been adapted successfully for image captioning - inputting an image and sampling the textual side of the model to generate a relevant caption~\cite{Zhou2020}. This begs the question: Can we go the other way and sample images from input pieces of text? i.e. \emph{Do vision-and-language BERT models know how to paint?}

In this work, we probe the ability of a powerful and popular representative from this family of models - \lxmert~\cite{Tan2019}, to produce high fidelity and semantically meaningful images conditioned on captions. Interestingly, our analysis leads us to the conclusion that \lxmert\ in its current form does not possess the ability to paint - it produces images that have little resemblance to natural images. This is a somewhat surprising finding given \lxmert's masked training objectives for both modalities and its impressive performance on tasks that seemingly require a similar skill set. 

We find that this is largely due to the regression training objective used by this family of models to predict masked features on the visual side. This is in contrast with the textual side, where they predict masked tokens within a large discrete vocabulary using a classification objective. Regressing features in high dimensional spaces is challenging to optimize and introduces noise at inference. This gets compounded when using iterative sampling procedures to predict the entire set of visual features. A downstream image generator consuming these predictions isn't able to recover from this noise even when fine-tuned on \lxmert's predictions.

We introduce \xlxmert\ that builds upon \lxmert\ and enables it to effectively perform discriminative as well as generative tasks. Our key refinements include: (a) simplifying the visual inputs to use grid features instead of object detection bounding boxes, (b) discretizing visual representations, (c) using uniform masking with a large range of masking ratios to enable the model to predict the entire set of visual clusters at inference time and (d) aligning the right pre-training datasets to the right objectives. When coupled with our proposed image generator, \xlxmert\ is able to generate rich imagery that is semantically consistent with the input captions. Importantly, \xlxmert's image generation capabilities rival state-of-the-art image generation models (designed only for generation), while its question answering capabilities show little degradation compared to \lxmert.

These refinements are not \lxmert-specific. They are designed to be easily applicable to a wide variety of multimodal BERT models. We find that \uniter, a single stream model for vision-and-language tasks, produces very poor images when coupled with a generator, but with our extensions, the resulting \xuniter\ produces images of a similar quality to \xlxmert.

In summary, we present \xlxmert, a unified multimodal transformer model that can answer questions, and also generate captions and images. Our extensions to enable these capabilities are not tied to \lxmert's underlying architecture. We expect that the entire family of multimodal BERT models can be enhanced with image generative capabilities using our introduced strategy.

\section{Related works}

\paragraph{Visual-Language transformer models}
Recent multi-modal pre-training models show significant improvements on a wide range of downstream tasks, including discriminiative (eg., visual question answering) and generation task (eg. image captioning \cite{Zhou2020}).
Some methods use a single transformer architecture to jointly encode text and image \cite{Li2019, su2019, Alberti2019, Rahman2019, Li2020, Chen2019, Qi2020, Huang2020}, while others use two-stream architectures \cite{Lu2019, Lu2020, Tan2019}.
These models typically consume object detection features. %
We probe this family of models at the task of image generation and present extensions that enable them to reliably generate images.

\paragraph{Sequence generation with undirectional transformer}
When generating sequences with conventional transformer language models, it is natural to sample tokens from left to right. However, since undirectional transformers (eg. BERT) are not trained with a specific generation order, a line of works has investigated different strategies for sequence generation with undirected models.
\citet{Wang2019} use Gibbs sampling from an all-mask sequence, and 
\citet{Dong2019, Bao2020a} use causal attention during training for left-to-right generation.
\citet{Liao2020, Mansimov2019, Ghazvininejad2019} sample masks from a uniform distribution during training for arbitrary order or parallel generation. We adapt these techniques for grid-based image generation.

\vspace{-.2cm}
\paragraph{Text-to-image synthesis}
Synthesizing images from text descriptions continues to be challenging. Since the pioneering work of \citet{Reed2016}, many methods have adopted GANs \cite{Goodfellow2014} to generate high-fidelity images. \citet{Nguyen2017} generate images that maximizes activation of a pretrained captioning model. Recent works \cite{Zhang2017, Zhang2018, Xu2018e, Li2019} use multi-stage generation, where low-resolution images are initially sampled, then gradually upsampled and improved in later stages. These models are specialized toward image generation, whereas our model can not just generate images, but also answer questions and generate captions. Also, our design is modular in nature. While we use a compact image generator with \xlxmert, one can also replace it with either of the aforementioned model architectures.

\vspace{-.2cm}
\paragraph{Grid visual representation}
Compared to bounding box representations which requires expensive object detection annotations, grid representations of images can be naturally obtained from CNNs. \citet{Jiang2020, Huang2020} have recently shown that these can be almost as powerful as bounding box representations for VQA.
Grid representation have been widely used in vision tasks, including self-supervised learning \cite{Oord2018, Henaff2019, Trinh2019, Gidaris2020, Noroozi2016} and image generation \cite{Oord2017, Lin2019b}. We leverage grid visual representations to enable \lxmert\ to generate images.

\section{Background: Revisiting LXMERT}
\label{sec:revisit_lxmert}

\begin{figure*}[t!]
\begin{center}
\includegraphics[
                width=\textwidth,
                 ]{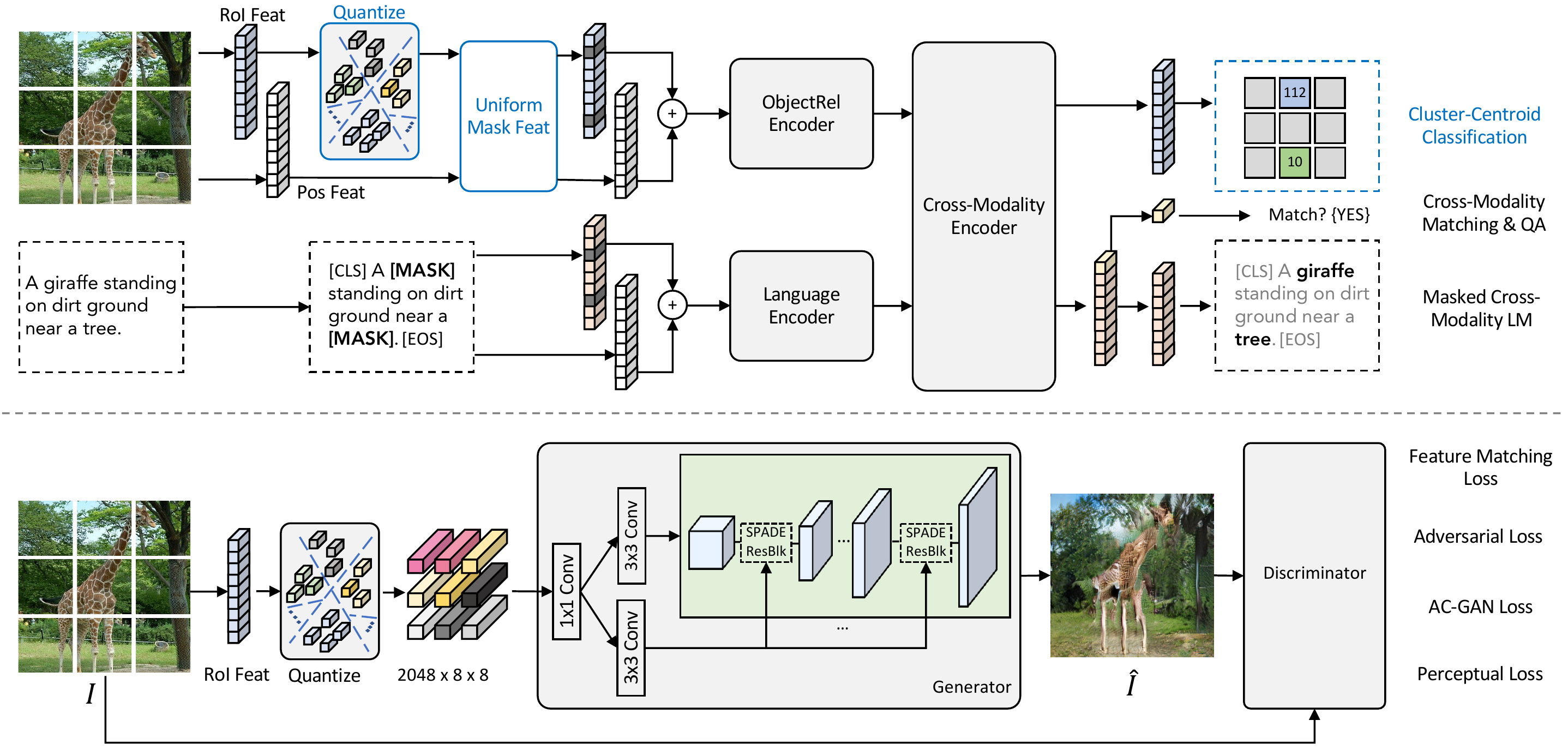}
\end{center}
\vspace{-.2cm}
\caption{
    \textbf{Top}: Overview of the proposed \xlxmert\ model. \textcolor[HTML]{1E90FF}{Blocks in blue} are the modifications we make to \lxmert\ model to enable it to paint. \textbf{Bottom}: Overview of the image generation architecture. The input to the model is a natural image that is compressed to a quantized latent map of size $8 \times 8$ by RoI Pooling. We use a generator consisting of multiple residual blocks with SPADE layer which encodes $8 \times 8$ grid features.
}
\vspace{-.3cm}
\label{fig:architecture}
\end{figure*}

Over the past year, a large number of transformer based architectures for multimodal data have produced impressive results across a variety of discriminative tasks. Some of these models have been shown to perform very well at the generative task of Image Captioning, but little attention has been paid to the reverse generative task: generating images given text. In this work, we first probe one popular representative from this family - \lxmert~\cite{Tan2019} - in its ability to paint; and propose extensions that enable it to paint.

\lxmert\ is a cross modality transformer with inputs: image $I$ and text $T$. This is represented as the sequence $\{v_1, \dots, v_\mathcal{T}, \texttt{CLS}, w_1, \dots, w_T, \texttt{EOS} \}$ where $\{v_i\}_{i=1}^\mathcal{T}$ are image region features, $\{w_j\}_{j=1}^T$ are word tokens and \texttt{CLS} and \texttt{EOS} are special tokens. \lxmert\ outputs embeddings for each input $\{h_{v_i}\}_{i=1}^\mathcal{T}$, $\{h_{w_j}\}_{j=1}^T$ and $h_\texttt{CLS}$, $h_\texttt{EOS}$. $h_\texttt{CLS}$ is used as the cross-modality output. Internally, \lxmert\ consists of two types of encoders: single-modality encoders for each modality and a cross-modality encoder using bi-directional cross attention to exchange information and align entities across the modalities. 

\lxmert\ is pretrained on several vision-and-language datasets with five objectives: 
Masked language modeling (MLM), Masked visual feature regression (MVFR) - reconstructing randomly masked words and regions given the remaining inputs, Masked object classification (MOC) - object classification on masked image regions, Image-text matching (ITM) - image-caption alignment prediction and Question answering (QA) - answering a question given image input. After pretraining, \lxmert\ is finetuned for various downstream tasks. Unless noted, we use the default settings and hyperparameters of \lxmert\ in our experiments.

\section{Probing LXMERT's Ability to Paint}

In order to probe \lxmert's ability to paint, we first modify its input image representation to a \textit{grid based} feature set (Sec.~\ref{sec:grid}) and then pass these to an image generator (Sec.~\ref{sec:gridgenerator}). 

\subsection{Grid Image Features} \label{sec:grid}
Most popular multimodal BERT models use image features extracted from the output of a Faster R-CNN \cite{Ren2015} object detector. The detected objects typically have various locations and sizes. Passing these features into an image generator poses some challenges:
(1) \lxmert\ is not trained to predict locations of given objects
(2) it is not trivial to predict both object classes and their locations simultaneously
(3) object detections do not cover backgrounds.

We modify \lxmert\ to use a uniform $N \times N$ grid and use RoI Pooling to extract the grid features. Note that we use the same detection backbone pretrained on the Visual Genome dataset to maintain parity with the original \lxmert. Our experiments in Sec~\ref{sec:experiment} show that moving to a grid based input causes very little degradation to downstream QA tasks, a finding consistent with \citet{Jiang2020}.

\vspace{-.2cm}
\paragraph{Sampling grid features:} Given text input, we  sample predicted visual features $\{h_{v_i}\}_{i=1}^\mathcal{T}$ where $\mathcal{T} = N \times N$ is the number of image regions, using Gibbs sampling in a manner similar to language generation using BERT by \citet{Wang2019}.

\subsection{Image Generation} \label{sec:gridgenerator} We use a compact image generator inspired by recent state of the art image synthesis methods leveraging Generative Adversarial Networks (GAN) \cite{Goodfellow2014}. Its takes as inputs an $N \times N$ grid of visual features from the pretrained Faster-RCNN network and generates an image. 
As shown in Fig~\ref{fig:architecture}, the input grid features are projected through convolutional layers and then passed to an image generator, which consists of multiple residual blocks \cite{Miyato2018}.
Each generator residual block has SPADE layer \cite{Park2019} which guides generator to outptut high fidelity images given semantic grid layouts.
In our experiments, we use an image generator which takes $8 \times 8$ grid features and outputs an $256 \times 256$ image.

\vspace{-.2cm}
\paragraph{Training the image generator:} The generator is pre-trained using $8 \times 8$ ground truth Faster-RCNN features, akin to \emph{teacher forcing}, without any inputs from \lxmert. 
We train the generator with the same loss as \citet{Park2019}, but replacing the segmentation map with a grid feature map. 

Fig.~\ref{fig:inference} (b) shows that our generation architecture can successfully reconstruct images using ground truth pre-trained grid features. Note that the generator still displays some reconstruction errors compared with modern auto-encoders such as VQ-VAEv2 \cite{Razavi2019} primarily due to (1) freezing the encoder backbone in order to match \lxmert's training settings (2) restricting grid features to have a low (and manageable) dimension.

\subsection{Can LXMERT Paint?} 
\begin{figure*}[t!]
\begin{center}
\includegraphics[
                width=\textwidth,
                 ]{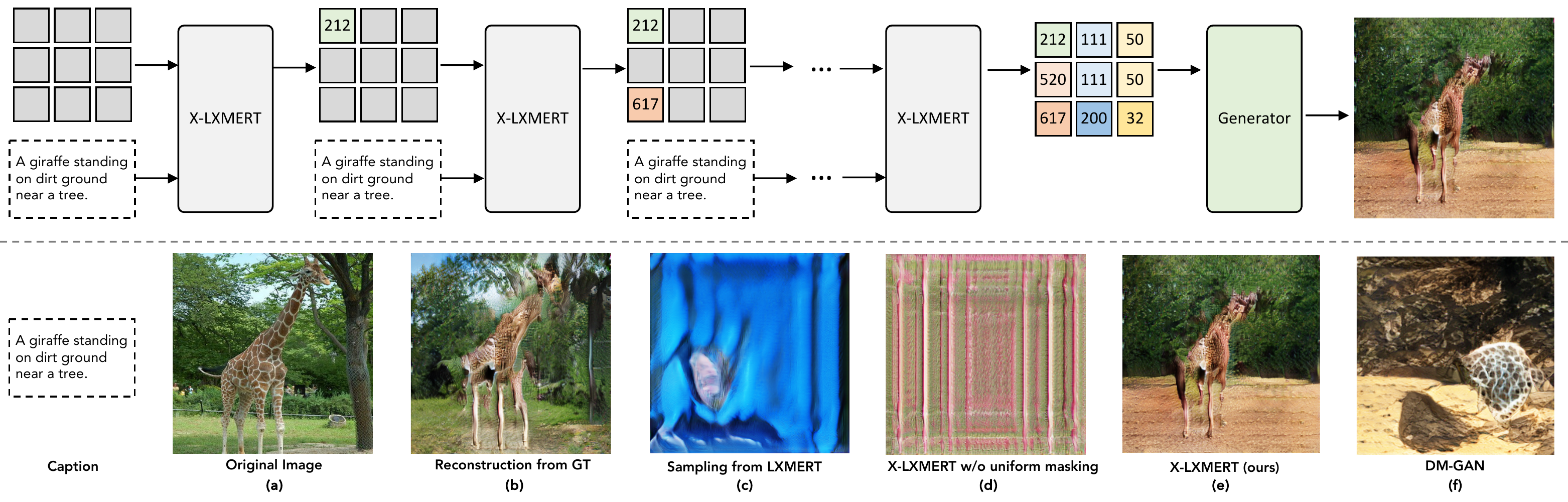}
\end{center}
\vspace{-.3cm}
\caption{
    \textbf{Top}: Image generation from \xlxmert. Given the text input and all masked visual feature, we first sample grid features by using Gibbs sampling with multiple iterations. Then the sampled grid features are fed into the generator to generate the image. 
    \textbf{Bottom}: Sampled images, from left to right (a) Original image (b) Reconstruction from GT features (c) Sampling from \lxmert\ + Grid (d) Sampling from \xlxmert\ without uniform masking pretraining (e) Our proposed \xlxmert\ (f) Generated image from DM-GAN \cite{Zhu2019}.
}
\vspace{-.3cm}
\label{fig:inference}
\end{figure*}

Our experiments in Section~\ref{sec:experiment} reveal that \lxmert\ is unable to produce visual features that can be converted to a meaningful image by a generator. Figure~\ref{fig:inference} shows an example.
Recall that the \lxmert\ loss function includes a regression loss - MVFR - that corresponds to regressing target visual features given the textual and visual context. Unfortunately, at inference, this loss on the validation set remains high, causing the predicted visual features to be fairly noisy. In addition, the Gibbs sampling procedure causes this error to propagate over the entire set of features. The resulting predictions aren't suitable to be used for downstream image generation.

\section{X-LXMERT}
In this section, we present \xlxmert\footnote{\xlxmert\ is an \lxmert\ with a ``display server''} that extends  \lxmert, enabling it to paint, while still maintaining a high performance on discriminative  tasks. \xlxmert\ has three key refinements that enable it to paint (Sec.~\ref{sec:refinements}): discretizing visual representations, using uniform masking with a large range of masking ratios, and aligning the right pretraining datasets to the right objectives. We then leverage Gibbs sampling to generate visual features given textual input (Sec.~\ref{sec:sampling}).

\subsection{From LXMERT to X-LXMERT}
\label{sec:refinements}
\noindent {\bf Discrete visual representations:} 
We observe that the visual features regressed by \lxmert\ are not suitable for image generation. Instead, akin to VideoBERT~\cite{Sun2019d}, we first create a visual vocabulary using K-mean clustering, approximate the target visual features via a nearest neighbor search, and modify \lxmert\ to predict the cluster ID for each masked visual token. A new Cluster-Centroid Classification objective (CCC) is used to replace the previous regression objective with a high cardinality classification objective. Our experiments show that discretizing visual representations results helps in predicting better visual features, stems the propagation of feature noise over sampling iterations and generates rich imagery.

\vspace{-.2cm} 
\paragraph{Uniform instead of Bernoulli masking:}  
Following BERT, \lxmert\  uses Bernouli sampling (with $p=0.15$) to determine  positions of the masked tokens on the visual and textual features. In order to generate an image from captions, all tokens on the vision side must be masked and predicted. A low probability Bernoulli sampling procedure does not prepare the model well for the generation task, and increasing the probability to very high values leads to poor pre-training. To resolve this, we use Uniform masking on the vision modality. \xlxmert's uniform masking first samples the masking ratio from a uniform prior distribution ([0,1]), and then samples the desired number of positions randomly. This subjects the model to a variety of masking ratios, and our experiments reveal that this greatly benefits image generation.

\vspace{-.2cm} 
\paragraph{Updating pre-training data:} \lxmert\ uses a variety of data to pre-train the model: QA data from multiple sources, caption data from \coco\ and captions from Visual Genome (VG).  Since \xlxmert\ uses the CCC loss function, predicting visual features given questions like: ``What is shown in the image?'' is very ambiguous and results in models that cannot predict visual clusters. Similarly, many captions from VG (e.g.,  ``A bag'' or ``Glasses on the hair'') tend to describe small regions of the image and not the whole image, which makes them unsuited to train the CCC objective. \xlxmert\ drops QA data and the captions from VG for CCC objective for visual cluster prediction. %

\subsection{Sampling Strategies for X-LXMERT}
\label{sec:sampling}

Given text input, predicting the entire set of visual features in one step does not produce good results. Instead, we employ Gibbs sampling to iteratively sample features at different spatial locations. In contrast to text generation, where left-to-right is considered a natural order, there is no natural order for generating images. The grid sampling process starts with $N^2$ grids filled with the \texttt{MASK} special token. The model then iteratively updates locations either one-by-one or multiple in parallel. There are several sampling strategies for sampling locations on the square grid, primarily falling into two buckets: auto-regressive and parallel.

\noindent \textbf{Autoregressive sampling}
In each iteration, a grid position is sampled, masked and predicted. Then the corresponding \texttt{MASK} token is replaced with the predicted one, and the process is repeated until all locations are updated. 

\begin{compactitem}[--]
\item TL$\rightarrow$BR: Positions are sequentially chosen from top-left to bottom-right, similar to PixelRNN \cite{Oord2016}.
\item Random \cite{Liao2020}: Positions are selected in random order. After $N^2$ steps, locations may be updated more than once.
\end{compactitem}

\noindent \textbf{Non-autoregressive sampling}
In each iteration, multiple positions are sampled, masked with \texttt{MASK}, predicted and then replaced.
\begin{compactitem}[--]
\item Mask-predict-K \cite{Ghazvininejad2019}: This requires K sampling steps. In the first iteration, all $N^2$ locations are updated. Then, we linearly decay the number of tokens updated per iteration. For example, for a $2 \times 2$ grid whereby $N^2 = 4$, if $K=4$ then (4, 3, 2, 1) positions are updated in each iteration. Within each iteration, positions with the lowest confidence are updated.
\end{compactitem}

\noindent Our experiments show that Mask-Predict-4 consistently produces good results across a variety of generation metrics and we propose using it for \xlxmert.
Our uniform masking aligns well with the linear decay of Mask-Predict and makes the model robust to a varied number of masked locations.

\subsection{Training Details}
\vspace{-.1cm}
\paragraph{Generator}
Following \cite{Park2019}, our generator and discriminator are jointly trained with 4 losses: (1) hinge adversarial loss \cite{Lim2017, Tran2017}, (2) AC-GAN loss \cite{Odena2017}, (3) discriminator feature matching loss \cite{Wang2018h} and (4) perceptual loss \cite{Johnson2016}. The coefficients for different loss are $(1,1,10,10)$ respectively. The perceptual loss is calculated with ResNet-50 \cite{He2016} pre-trained on ImageNet \cite{Deng2009}.  
We use Adam optimizer \cite{Kingma2015} with $(\beta^1, \beta^2)=(0, 0.999)$ and two-time update rule \cite{Heusel2017} with learning rate of 0.0004 and 0.0001 for generator and discriminator respectively.
We train the generator with batch size 96 for 60 epochs.
Note that the generator parameters are fixed after training and not finetuned.
Please refer Sec.~\ref{sec:generator_details} for more details.

\vspace{-.2cm}
\paragraph{Pre-training}
Following \lxmert\ \cite{Tan2019},
we use AdamW optimizer \cite{Loshchilov2019} with $(\beta^1, \beta^2)=(0.9, 0.999)$ and learning rate \texttt{1e-5} with 5\% linear warmup schedule.
We train \xlxmert\ on with batch size 920 for 20 epochs.
Instead of using all pretraining tasks for each step, we first uniformly sample a modality to mask from \texttt{[image, text, no-mask]} and run corresponding tasks.
Please refer to Sec.~\ref{sec:pretraining_detail} for more details.

\vspace{-.2cm}
\paragraph{Finetuning}
For each downstream task, a task head consisting of two fully connected layers is trained along with pre-trained \xlxmert. We used the same parameter setting with \lxmert.
Please refer to Sec.~\ref{sec:finetuning_detail} for more details.

\section{Experimental Setup}
\label{sec:experiment}

In this section we present experimental setups to evaluate image generation, visual question answering and visual reasoning.

\subsection{Evaluating Image Generation}
\label{sec:eval_image_generation}

We train and evaluate models using the MS \coco\ captioning dataset~\cite{Lin2014}. We compare \xlxmert\ with \lxmert\ and state-of-the-art text-to-image generation methods: StackGAN \cite{Zhang2018}, PPGN \cite{Nguyen2017}, AttnGAN \cite{Xu2018e}, ControlGAN \cite{Li2019}, and DM-GAN \cite{Zhu2019}.
Image generation is a particularly difficult task to evaluate, due to the variability in acceptable outputs for a given caption, as well as the subjective nature of perceiving image quality. We present a suite of automated and manual metrics to compare models.
\newcommand{\mm}{\cellcolor{blue!10}}
\newcommand{\g}{\cellcolor{green!10}}

\begin{table*}[!t]\footnotesize
  \centering
\resizebox{\textwidth}{!}{
\begin{small}
\begin{tabular}{l cc c c c c c ccccc}
\toprule
    \multicolumn{1}{c}{} & \multicolumn{7}{c}{{Text-to-Image Generation}} & \multicolumn{3}{c}{{Visual Question Answering}} & \multicolumn{2}{c}{{Visual Reasoning}}\\
    \cmidrule(lr){2-8}
    \cmidrule(lr){9-11}
    \cmidrule(lr){12-13}
     \multirow{2}{*}{Methods} & \multirow{2}{*}{IS$\uparrow$} & \multirow{2}{*}{FID$\downarrow$} & \multirow{2}{*}{\shortstack{R-prec\\-easy$\uparrow$}} & \multirow{2}{*}{\shortstack{R-prec\\-hard$\uparrow$}} & 
    \multirow{2}{*}{\hummus} & 
    \multicolumn{2}{c}{Human pairwise pref}  & 
    \multicolumn{2}{c}{VQA} & \multicolumn{1}{c}{GQA} & \multicolumn{2}{c}{NLVR\textsuperscript{2}} \\
    \cmidrule(lr){7-8}
    \cmidrule(lr){9-10}
    \cmidrule(lr){11-11}
    \cmidrule(lr){12-13} 
    & & & & & & Semantics & Fidelity & test-dev & test-std & test-std & dev & test-P \\
    \midrule 
    Original Image                 & 36.6 & - & 89.6 & 47.6  & 0.73 & - & - & - & - & - & - & - \\
    \midrule
    StackGAN               & 8.5 & - & - & - & - & - & - & - & - & - & - & - \\
    PPGN                  & 9.6 & - & - & - & - & - & - & - & - & - & - & - \\
    AttnGAN                  & 25.9 & 35.5 & - & - & - & - & - & - & - & - & - & - \\
    ControlGAN                & 24.1 & - & - & - & - & - & - & - & - & - & - & - \\ 
    \hhline{~~~~~~--}
    DM-GAN          & \textbf{30.5} & \textbf{32.6} & \textbf{51.8} & \textbf{27.5} & \textbf{0.49} & \multicolumn{1}{|c}{\g 37.0} & \multicolumn{1}{|c|}{\mm 35.9} & - & - & - & - & -\\ 
    \hhline{>{\arrayrulecolor{black}}------>{\arrayrulecolor{green!10}}->{\arrayrulecolor{blue!10}}->{\arrayrulecolor{black}}-----}
    \bfseries{\scshape{X-Lxmert}} & 22.7 & 37.4 & 40.8  & 25.1  & \textbf{0.49} & \multicolumn{1}{|c}{\textbf{\g 52.0}} & \multicolumn{1}{|c|}{\textbf{\mm 50.0}} & 68.6 & 68.7 & 58.4 & 72.4 & 72.4  \\ \cline{7-8}
    \lxmert*+Grid  & 1.6 & 316.7 & 0.5 & 6.6 & 0.27 & & & 71.1 & 71.2 & 60.1 & 74.6 & 74.0 \\
    \midrule
    \lxmert  & - & - & - & - & -  & - & - & \textbf{72.4} & \textbf{72.5}  & \textbf{60.3} & \textbf{74.9} & 74.5 \\
    \lxmert* & - & - & - & -  & - & - & - & 70.9 & 71.1  & 59.9 & \textbf{74.9}  & \textbf{75.0}  \\
    
    \bottomrule
  \end{tabular}
  \end{small}}
  \caption{Comparing \xlxmert, \lxmert\ and baselines on image generation, visual question answering and visual reasoning tasks.  The pairwise metric compares \lxmert\ and DM-GAN; numbers do not sum to 100 due to the TIE option provided to annotators.  Note that \xlxmert\ and \lxmert*+Grid are the only models that are able to produce results for all tasks. *: Our re-implementation of \lxmert. }
  \label{table:image_metrics}
  \vspace{-10pt}
\end{table*}

\vspace{-.2cm}
\paragraph{Automated Metrics: Evaluate image quality} We use Inception score (IS) \cite{Salimans2016a} to measure image diversity and Fréchet Inception Distance (FID) \cite{Heusel2017} to measure authenticity; using Inception v3 \cite{Szegedy2016} as a surrogate net.

 \vspace{-.2cm}
\paragraph{Automated Metrics: Evaluate semantics}
We use two variants of R-precision \cite{Xu2018e}, R-prec-easy and R-prec-hard to evaluate if the image is well conditioned on the input text. Given a generated image, a positive caption and negatives, R-precision measures the retrieval rate for the positive caption using a surrogate multi-modal network. We use an independent surrogate - ViLBERT-MT \cite{Lu2020} for this purpose. R-prec-easy is the variant of R-precision with easy negatives (sampled randomly amongst the caption set). R-prec-hard is the variant with hard negatives (swapping a word in a caption with another word within the same category, e.g., red $\Rightarrow$ green). We choose words from one of 4 categories: nouns (80 \coco\ objects), 64 verbs, 10 colors and 10 numbers.

The above automatic metrics, while cheap and reproducible, are noisy because they depend on imperfect surrogate models. The ultimate measure of quality and semantics for image generation continues to be crowd-sourced human studies. 

 \vspace{-.2cm}
\paragraph{Human Study: Pairwise preferences}
We conduct a human preference evaluations between \xlxmert\ and the best performing model in the automated metrics---DM-GAN. We measure (1) \emph{Semantic preference} by showing two image and asking annotators to select the one that best matches the source caption. (2) \emph{Fidelity preference} by showing the two images alone and asking which appears more realistic. Both evaluations also allow a third option (Tie) to be selected. For each evaluation, 5000 image pairs were used, and 357 unique crowdworkers participated in total (median annotations per worker---17). 

 \vspace{-.2cm}
\paragraph{Human Study: Our new metric -- HUMMUS}
The above pairwise test is very useful and widely used to evaluate generative models, but measuring new models becomes challenging, since they must compare to all old models. To expand human evaluation, 
we present a novel metric to test semantic consistency between the caption and image inspired by masked token modeling, named - HUmans Measuring seMantics Using maSking (\hummus). To compute \hummus, human annotators are shown an image and its caption with a single word masked out. They are asked to complete the partial caption based on information in the image, and a match is counted only when a majority of annotators supply the correct word. The total score is reported as a ratio of these successful matches. The task was run on 2800 image-caption pairs (2289 unique images), with 5 annotators per pair. A total of 280 unique crowdworkers completed the task, with a median of 13 images annotated per worker.  A high \hummus\ score reveals that the generated images contain the corresponding semantics, well enough to be recognized. The masked word is chosen from one of 3 categories: 80 \coco\ nouns, verbs and colors.

\subsection{Evaluating Visual Question Answering}
We train and evaluate models for visual question answering using the VQA2.0~\cite{Goyal2019} and GQA~\cite{Hudson2019} datasets, which provide an image and a question and require the model to generate an answer.

\subsection{Evaluating Visual Reasoning}
We train and evaluate models for visual reasoning using the \nlvr~\cite{Suhr2019} dataset and report numbers on the dev and test-P splits. The \nlvr\ dataset requires models to look at two images and determine if an accompanying caption is True or False. This is a particularly challenging dataset for present day vision and language models.

\section{Experimental Results}
\label{sec:results}

We now present a comparison of \xlxmert with several baselines on the generative and discriminative tasks, along with ablation studies and qualitative results. We also show the generality of our techniques via extending \uniter\ to create \xuniter.

\subsection{Quantitative Results}
Table~\ref{table:image_metrics} provides detailed metrics for \xlxmert\ and baselines. It also provides generation metrics for the original image in the dataset for the corresponding input text. Note that \xlxmert\ and \lxmert+Grid are the only models that are able to produce results for all tasks.

\begin{figure*}[h]
\begin{center}
\includegraphics[
                width=\textwidth,
                 ]{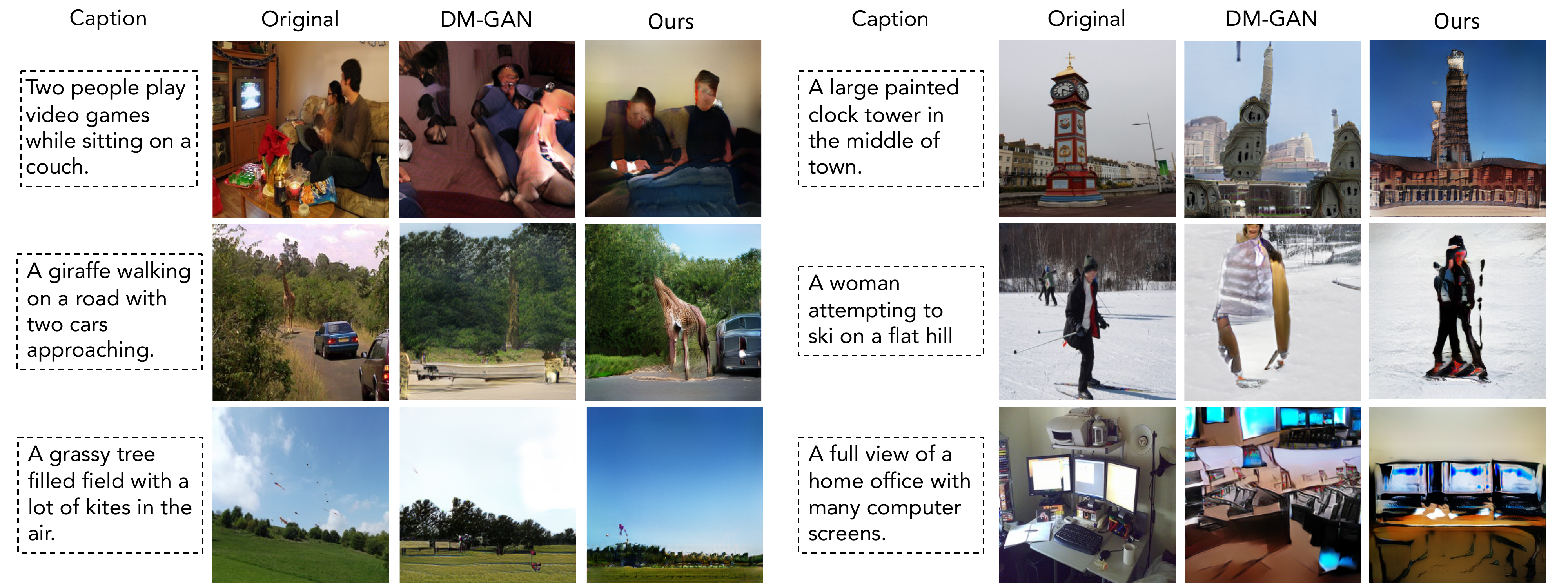}
\end{center}
\vspace{-10pt}
\caption{Qualitative examples of input caption, original images, image generated by DM-GAN\footnote{We use coco pre-trained model from https://github.com/MinfengZhu/DM-GAN} \cite{Zhu2019} and images generated by our proposed \xlxmert.}
\vspace{-12pt}
\label{fig:image_generation_qualitative}
\end{figure*}

 \vspace{-.2cm}
\paragraph{Image Generation}
As seen, \xlxmert\ significantly outperforms \lxmert\ across all generation metrics. \xlxmert\ even outperforms two specialized generation models, comparable to AttnGAN and ControlGAN. 
Our model is lower compared to DM-GAN in terms of automated metric (IS and FID), however, it is competitive with DM-GAN at semantic metric (R-prec-hard)\footnote{Note: R-prec and HUMMUS are reported only for DM-GAN (the strongest of the 5 baselines), since this was the only model with code and pretrained weights. IS and FID numbers are from their respective publications or from \citet{Zhu2019}. The detailed R-prec-hard numbers across categories are presented in the appendix.}.

Note that \xlxmert's image generator is much smaller than the one used by DM-GAN (1.7M vs 22.3M parameters). While the transformer employed in \xlxmert\ is large, it is a unified textual and visual encoder used for multiple tasks and is not finetuned for image generation. We expect \xlxmert's image quality to improve further when coupled with a larger image generator such as the one by DM-GAN. 

Table~\ref{table:image_metrics} also presents \hummus\ scores. Here we see that the semantics generated by \xlxmert\ is on par with DM-GAN and still significantly better than \lxmert. All models are still a distance away from the original image. \hummus\ matches on the lemmatized forms of masked words to allow for lexical variation, but it misses synonyms and other valid descriptors. This causes the score for the original image to drop to its reported value. See the appendix for R-prec-hard and HUMMUS broken down into categories.

Finally we present human pairwise preference scores between \xlxmert\ and DM-GAN (its closest competitor). Here we see that human annotators clearly prefer \xlxmert\ to DM-GAN for semantics as well as fidelity.

In summary, \xlxmert's generation capabilities rival state of the art specialized generation models. In fact, our human studies demonstrate that \xlxmert\ produces better results than even DM-GAN, its closest competitor. Our analysis also shows the limitations of current automatic evaluation metric for text-to-image synthesis.

\vspace{-.2cm}
\paragraph{Visual Question Answering}
Table~\ref{table:image_metrics} compares models on the VQA2.0 and GQA datasets. Converting \lxmert\ to use grid inputs causes a slight or no drop, consistent with findings by \citet{Jiang2020}, but hugely simplifies the pipeline. \xlxmert\ shows 1.5 - 2.5\% drop on these datasets but note that its numbers are still very competitive.

\paragraph{Visual Reasoning}
Table~\ref{table:image_metrics} compares models on \nlvr\ dataset. Consistent with VQA, grid inputs cause a slight drop. \xlxmert\ shows a roughly 2\% drop but retains most of the massive jumps obtained by \lxmert\ on \nlvr\ compared to the previous generation of models. 

Our implementation of \xlxmert\ uses a small $8 \times 8$ grid. Increasing the grid size will likely shrink gaps in VQA2.0, GQA and \nlvr\ datasets as per the recent findings by \citet{Jiang2020}.

\subsection{From \xlxmert~to \xuniter}
The proposed refinements (Sec.~\ref{sec:refinements}) to enable image generation capabilities are not \lxmert-specific. We apply these changes to \uniter~\cite{Chen2019}, a single stream multi-modal transformer architecture.
Instead of following \cite{Chen2019}
Table~\ref{table:ablation_uniter} shows that \uniter\ + Grid produces very poor images, but \xuniter\ obtains image generation scores comparable to \xlxmert -- showing the generality of our extensions.

\begin{table}[!h]\footnotesize
  \setlength\tabcolsep{12 pt}
  \centering
  \begin{tabular}{lcc}
    \toprule
     & IS$\uparrow$ & FID$\downarrow$ \\
    \midrule
    \uniter\ + Grid  & 2.4 & 253.5 \\
    \xuniter\        & {\bf 20.1} & {\bf 51.4} \\ \hline
    \lxmert\ + Grid  & 1.6 & 316.7 \\
    \xlxmert\        & {\bf 22.7} & {\bf 37.4} \\ 
    \bottomrule
  \end{tabular}
  \caption{Adding image generation capabilities to \lxmert\ and \uniter.}
  \label{table:ablation_uniter}
  \vspace{-10pt}
\end{table}

\subsection{Qualitative Results}

\begin{figure*}[h!]
\begin{center}
\includegraphics[
                width=\textwidth,
                 ]{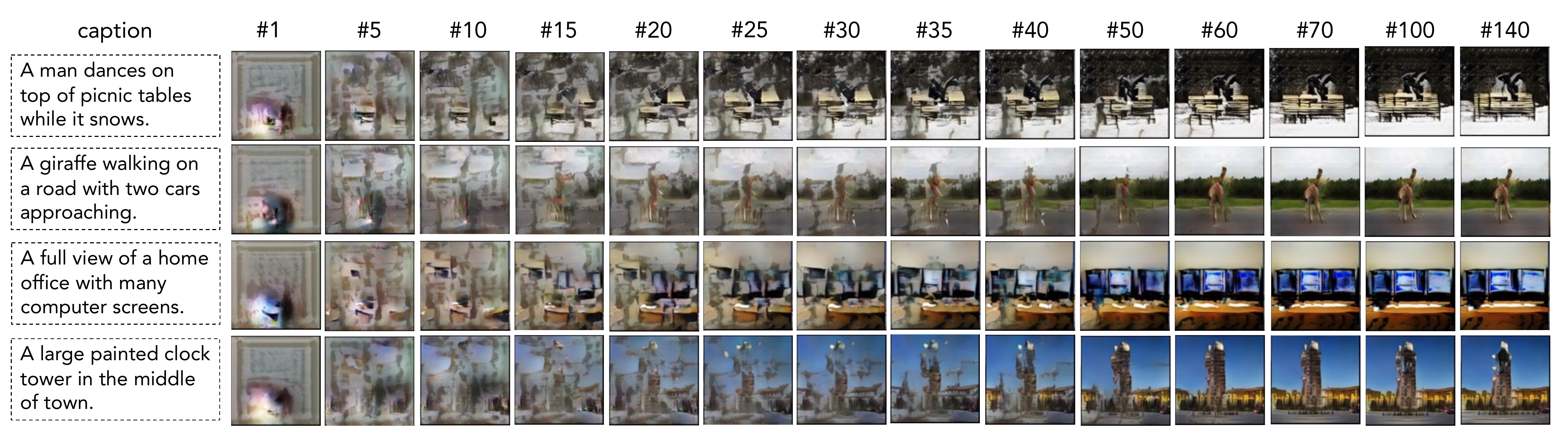}
\end{center}
\vspace{-13pt}
\caption{Intermediate images generated by \xlxmert\ at during 140 steps of random position sampling. Images are gradually improved as sampling steps proceed.}
\label{fig:intermediate_sample}
\end{figure*}

\begin{figure*}[t!]
\begin{center}
\includegraphics[
                width=\textwidth,
                 ]{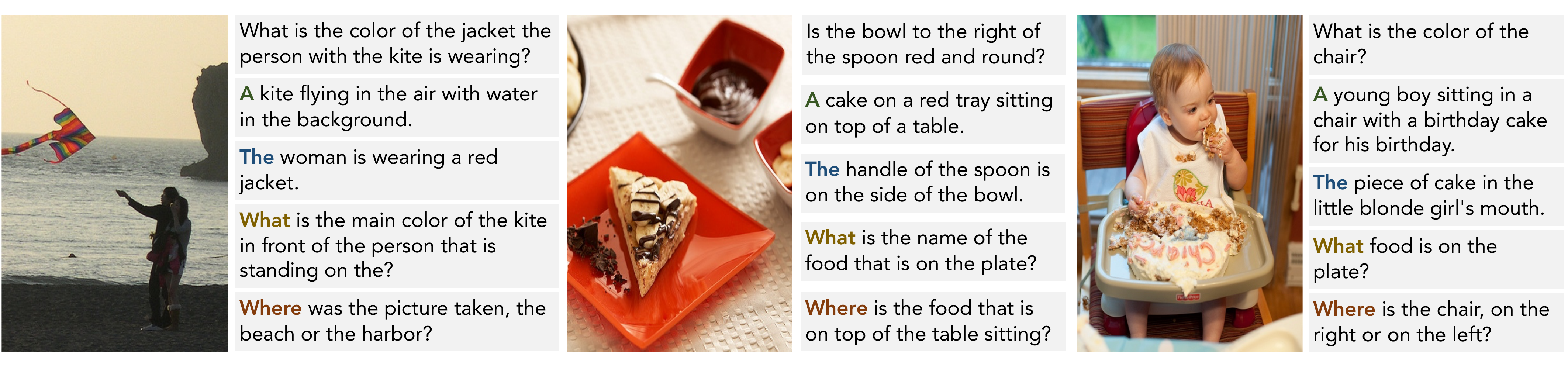}
\end{center}
\vspace{-11pt}
\caption{Captions generated by \xlxmert\ using Gibbs sampling. We control the samples by providing different prefix word into the model. Those prefix words are common starting word such as `\emph{A}', `\emph{The}', `\emph{What}', `\emph{Where}'.}
\vspace{-10pt}
\label{fig:captions}
\end{figure*}

\begin{table}[t]\footnotesize
  \centering
  \begin{tabular}{lccc}
    \toprule
    Ablations & IS$\uparrow$ & FID$\downarrow$ \\
    \midrule
    \lxmert\ + Grid & 1.6 & 316.7 \\
    \bfseries{\scshape{X-Lxmert}} & \textbf{22.7} & \textbf{37.4} \\
    \quad w/o discrete visual representations & 1.5 & 304.4 \\
    \quad w/o uniform masking & 2.1 & 227.9 \\
    \quad w/o updating pre-training data & 21.6 & 46.1 \\
    
    \bottomrule
  \end{tabular}
  \caption{An ablation study for the three proposed refinements.}
  \label{table:ablation}
  \vspace{-10pt}
\end{table}

Fig~\ref{fig:image_generation_qualitative} shows qualitative examples by \xlxmert\ compared to DM-GAN \cite{Zhu2019}. While the images lack fine details, they do a reasonable job at preserving high level semantics, as revealed by the metrics. For complex scene, our model is able to preserve better semantics (e.g. `two people', `clock tower` and `home office') compared to DM-GAN. We do not show images produced by \lxmert\ since they largely tend to be incomprehensible.

To better understand the image generation process, We show intermediate images generate by \xlxmert in Fig~\ref{fig:intermediate_sample}. We use random autoregressive sampling with 140 steps. Interestingly, the model first coarsely generates salient objects (ex. giraffe, monitors) in the caption followed by details and background.
 
Our model is able to generate captions given image. For each image, we sample text from \xlxmert\ using Gibbs sampling as shown in Fig~\ref{fig:captions}. We control the samples by providing different prefix word into the model. Those prefix words are common starting word such as `A', `The', `What', `Where'. \xlxmert\ can produce long meaningful captions as well as questions (like the ones in VQA datasets).

\subsection{Ablation Studies}
We examine the effects of our proposed refinements and our sampling strategies to the image generation quality. Table~\ref{table:ablation} shows that two of the proposed refinements to \lxmert\ (moving to discrete visual representations and using uniform masking) are critical to produce high quality images. The third refinement -- updating pre-training data for the CCC objective -- is less critical, but useful nonetheless.

\begin{table}[t]
  \centering
\resizebox{\columnwidth}{!}{
\begin{small}

    \begin{tabular}{l ccc }
    \toprule
     & IS$\uparrow$/FID$\downarrow$ & R-prec$\uparrow$  & \hummus\ $\uparrow$\\
     & & easy/hard & Noun / Verb / Color / Avg.\\
    \midrule
    \textbf{Mask-Pred-4} & \textbf{22.7}/37.4 & \textbf{40.8}/\textbf{25.1} & \textbf{0.55} / \textbf{0.42} / 0.50 / \textbf{0.49} \\
   TL$\rightarrow$BR & 19.8/48.5 & 26.9/18.9  & 0.45 / \textbf{0.42} / 0.41 / 0.43\\
    Random & 22.6/\textbf{35.9} & 39.5/24.7         & 0.52 / \textbf{0.42} / 0.51 / 0.48\\
    Mask-Pred-1 & 19.5/51.4 & 36.8/21.4           & 0.48 / 0.40 / \textbf{0.54} / 0.47  \\
    \bottomrule
  \end{tabular}
  \end{small}}
  \caption{Comparing image quality across sampling strategies. We propose to use Mask-Predict-4 for the default sampling strategy.}
  \label{table:image_ablation_sampling}
  \vspace{-10pt}
\end{table}

Table~\ref{table:image_ablation_sampling} shows that \xlxmert\ is fairly robust to sampling strategy, particularly for image semantics, with the exception of TL$\rightarrow$BR which tends to produce worse results. This is interesting in that TL$\rightarrow$BR is typically the default strategy used by practitioners \cite{Oord2016, Oord2017}. However, note that the differences between the strategies are quite small.

\section{Conclusion}
We develop a probing mechanism and find that \lxmert, a powerful   vision-and-language transformer model, is not able to generate meaningful images conditioned on text. We present \xlxmert, a unified model for image generation, captioning, QA and visual reasoning, and show that our extensions can easily be applied to other vision-and-language transformer models.

\newpage
\bibliographystyle{acl_natbib}
\bibliography{references}

\raggedbottom

\pagebreak
\clearpage
\appendix

\section{Qualitative samples}

\begin{figure*}[t]
\begin{center}
\includegraphics[
                width=\textwidth,
                 ]{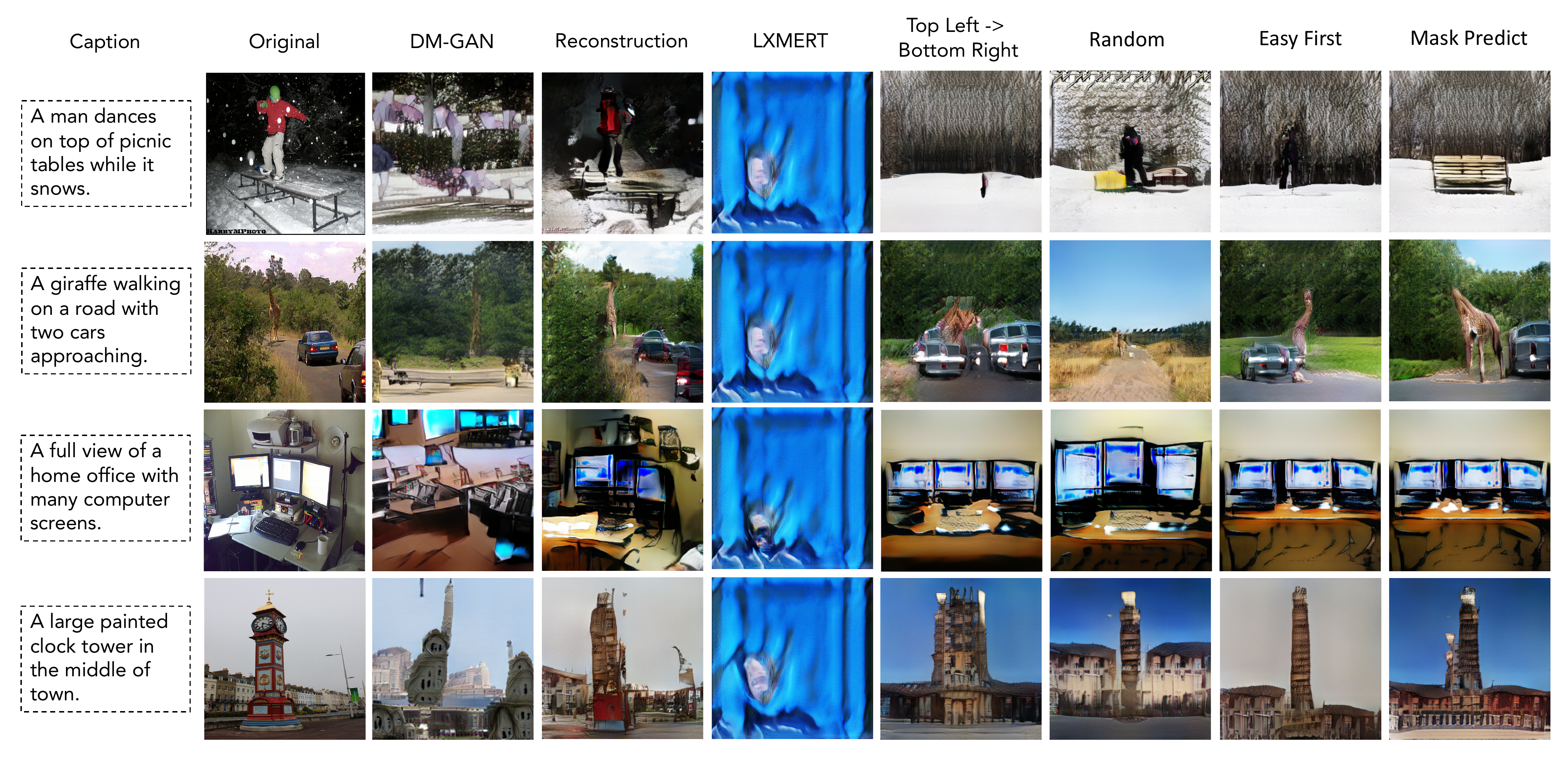}
\end{center}
\vspace{-10pt}
\caption{More qualitative examples of images generated by \xlxmert.}
\label{fig:image_generation_qualitative_more}
\end{figure*}

\begin{figure*}[!h]
\begin{center}
\includegraphics[
                width=\textwidth,
                 ]{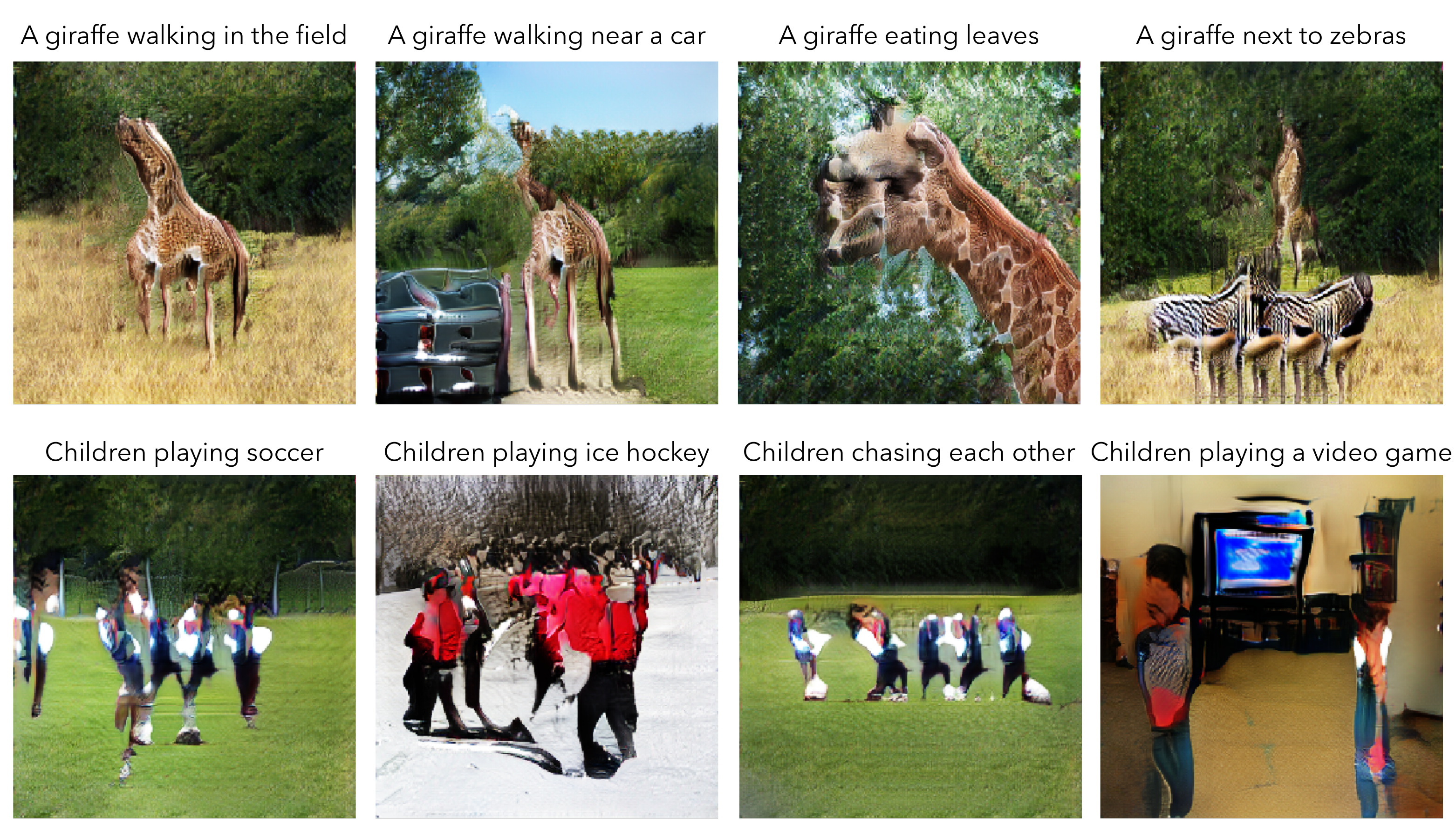}
\end{center}
\vspace{-10pt}
\caption{Images generated by \xlxmert\ demonstrating its ability to place objects within varied contexts.}
\label{fig:image_generation_qualitative_2}
\end{figure*}

\paragraph{More qualitative samples}
In Fig~\ref{fig:image_generation_qualitative_more}, we show more qualitative examples of images generated by DM-GAN, reconstruction from ground truth clusters, \lxmert, our proposed \xlxmert\ with different sampling strategies. Fig~\ref{fig:image_generation_qualitative_2} shows images generated by \xlxmert\ with the same subject placed in a variety of contexts.

\section{Source code}
Please refer to the project page for more details about this research, at \url{https://prior.allenai.org/projects/x-lxmert}.
This includes an animation of the iterative image generation process, a demo of \xlxmert\ accessible at \url{https://vision-explorer.allenai.org/text_to_image_generation} and code available at \url{https://github.com/allenai/x-lxmert}.

\section{LXMERT / X-LXMERT details}

For a fair comparison, we re-implement \lxmert\ and \lxmert\ with grid features.   
Our models have 226.5M trainable parameters, slightly smaller than 228M of original \lxmert\ implementation due to weight sharing of MVFR head and MOC head.
We use PyTorch \cite{Paszke2017} and Huggingface Transformers \cite{Wolf2019} libraries for implementation.

\subsection{LXMERT Architecture}
\lxmert\ architecture consists of text embedder, object embedder, transformer backbone, and task-specific heads.

\paragraph{Text embedder}
A text input is tokenized by WordPiece Tokenizer \cite{Wu2016a} and special tokens \texttt{CLS} and \texttt{EOS} are concatenated: $\{\texttt{CLS}, w_1, \dots, w_T, \texttt{EOS}\}$.
We use the same vocabulary used in BERT\footnote{\texttt{bert-base-uncased}} and \lxmert\ with size 30522.
Text is truncated with maximum token length of 20, including two special tokens.
768-dimensional embedding is learned for each token and position.
Final text embedding is obtained by sum of token embedding and positional embedding.

\paragraph{Object embedder}
An input image is resized within minimum length 800 and maximum length 1333 while preserving aspect ratio.
We use Faster R-CNN trained on Visual Genome to extract 36 bounding boxes from each image\footnote{We use PyTorch version (\url{https://gitlab.com/vedanuj/vqa-maskrcnn-benchmark}), instead of Caffe version (\url{https://github.com/peteanderson80/bottom-up-attention}) used in original implementation.}.
We take \texttt{fc6} feature, which is between \texttt{RoI-Pool} layer and final object classification head and has 2048 dimension. This is encoded into 768 dimensional vector followed by layer norm \cite{Ba2016}.
Four bounding box coordinates $(x_0, x_1, y_0, y_1)$ are $[0, 1]$-normalized by width and height. Then they are also encoded into 768 dimensional vectors with fully connected layer followed by layer norm.
Final object embedding is obtained by element-wise average of object and positional feature.

\paragraph{Transformer backbone}
Transformer backbone of \lxmert\ consists of object relation encoder, language encoder and cross modality encoder, which are composed of 9 self-attention layer \cite{Vaswani2017}, 5 self-attention layer, and 5 cross-attention layer respectively. The self-attention layers are same as the ones used in BERT and the dimension of the layers is 768.

\paragraph{Task-specific heads}
\lxmert\ is pretrained with five objectives\footnote{We do not use 400 object attributes predicted from Faster R-CNN, which were used by original implementation.} (MLM, MVFR, MOC, ITM, QA) as explained in Sec. \ref{sec:revisit_lxmert}.
For MLM, MVFR, ITM, QA task, a task head consisting of two fully connected layers with GeLU activation \cite{Hendrycks2016} and layer norm is trained.
For MOC task, a fully connected layer is applied on ouput of MVFR head, similar to original object detection pipeline\footnote{Original implementation trains separate head for MOC task.}.
For MLM, MVFR, MOC tasks, task heads are applied on cross-modal encoder outputs corresponding to masked tokens.
For ITM, QA tasks, tasks heads are applied on \texttt{CLS} token.

\subsection{X-LXMERT Architecture}

\xlxmert\ shares most components with \lxmert, except for minor modifications below.

\paragraph{Object embedder $\rightarrow$ Grid embedder}
We extract $8 \times 8$ grid features of \texttt{fc6} layer of Faster R-CNN, by giving positional information of  $8 \times 8$ grids into \texttt{RoI-Pool} layer. Then we quantize these features with nearest neighborhood search from 10,000 cluster centroids. Remaining are same with object embedder of \lxmert.

\paragraph{MOC, MVFR tasks $\rightarrow$ CCC task}
We replace MOC, MVFR tasks with CCC task (see Sec. \ref{sec:refinements}) for \xlxmert.
For CCC head, we simply modify the output dimension of fully connected layer used in MOC task to the number of clusters (1600 $\rightarrow$ 10000).

\subsection{Datasets}
For pretraining, we use same datasets used in \lxmert.
We use vision-and-language datasets whose images come from MS \coco~\cite{Lin2014} or Visual Genome~\cite{Krishna2016}.
Besides the two original captioning datasets, we also aggregate three large image question answering (image QA) datasets: VQA v2.0~\cite{Goyal2019}, GQA balanced version~\cite{Hudson2019}, and VG-QA~\cite{Zhu2016}.
Table \ref{table:lxmert_data} shows statistics of the datasets.
Note that \xlxmert\ only uses \coco\ captions for CCC task.

\begin{table*}[]
\centering
\begin{tabular}{cccccccc}
\toprule
\multirow{2}{*}{Image Split} & 
\multirow{2}{*}{Images}
& \multicolumn{6}{c}{Sentences (or Questions) }\\
\cmidrule(lr){3-8}
            &  & COCO-Cap & VG-Cap & VQA  & GQA & VG-QA & All   \\
\midrule
MS COCO - VG & 72K & 361K  & - & 387K & - & - & 0.75M  \\
MS COCO $\cap$ VG & 51K & 256K & 2.54M & 271K & 515K  & 724K  & 4.30M \\
VG - MS COCO & 57K  & -   & 2.85M  & - & 556K  & 718K  & 4.13M \\
\midrule
All & 180K & 617K & 5.39M  & 658K & 1.07M & 1.44M & 9.18M \\
\bottomrule
\end{tabular}
\caption{Dataset statistics used in pretraining. Each image has multiple sentences/questions. `Cap' is caption. `VG' is Visual Genome. Since MS COCO and VG share $51$K images, we list it separately to ensure disjoint image splits. This table is from \lxmert\ \cite{Tan2019}.}
\vspace{-8pt}
\label{table:lxmert_data}
\end{table*}

\subsection{Visual vocabulary clustering}
To create visual vocabularies, we run K-means clustering on Faster R-CNN grid features of \coco\ \texttt{train2014} images.
\texttt{train2014} has 82783 images, resulting 8 x 8 x 82783 = 5.3M grid features.
We use FAISS \cite{Johnson2017} library for clustering.
We sample 2.6M features in training data and run 20 iteration, which takes 2 hours.

\subsection{Training}
\label{sec:pretraining_detail}
We train \lxmert\ and \xlxmert\ for 20 epochs with mixed precision using Apex\footnote{\url{https://github.com/NVIDIA/apex}} (opt-level \texttt{O1}).
We use AdamW optimizer \cite{Loshchilov2019} with $(\beta^1, \beta^2)=(0.9, 0.999)$ and learning rate \texttt{1e-5} with 5\% linear warmup schedule.
We use gradient clipping with maximum norm 1.

Instead of using all pretraining tasks for each step, we first uniformly sample a modality to mask from \texttt{[image, text, no-mask]} and run corresponding tasks similar to \cite{Chen2019, Lu2020}.
When \texttt{image} is selected, we use MVFR, MOC for \lxmert\ and CCC for \xlxmert.
When \texttt{text} is selected, we use MLM.
When \texttt{no-mask} is selected, we replace given text with a random sentence from training data with 0.5 probability. If the text is replaced, we use ITM. If not, we use ITM and QA.

Training \lxmert\ takes 60 hours with batch size 1280, and training \xlxmert\ takes 40 hours with batch size 920.
We use 4 Titan RTX GPUs  ($4 \times 24$GB) for training both models.

\subsection{Finetuning}
\label{sec:finetuning_detail}
During finetuning on VQA/GQA/\nlvr, a task head consisting of two fully connected layers with GeLU activation and layer norm is trained along with pre-trained \lxmert\ and \xlxmert. For VQA/GQA, the parameters are initialized from pretrained QA head.
We use AdamW optimizer with learning rate \texttt{5e-4}.
We train \lxmert\ and \xlxmert\ for 10 epochs for each task.
For VQA/GQA/NLVR\textsuperscript{2}, finetuning takes 3/5/1 hours respectively on 4 Titan RTX GPUs  ($4 \times 24$GB).

\section{Generator details}
\label{sec:generator_details}

Our image generation system adopts GAN \cite{Goodfellow2014} framework and has two networks trained: generator and discriminator.

\subsection{Generator Architecture}

\begin{figure*}[h!]
\begin{center}
\includegraphics[
                width=\textwidth,
                 ]{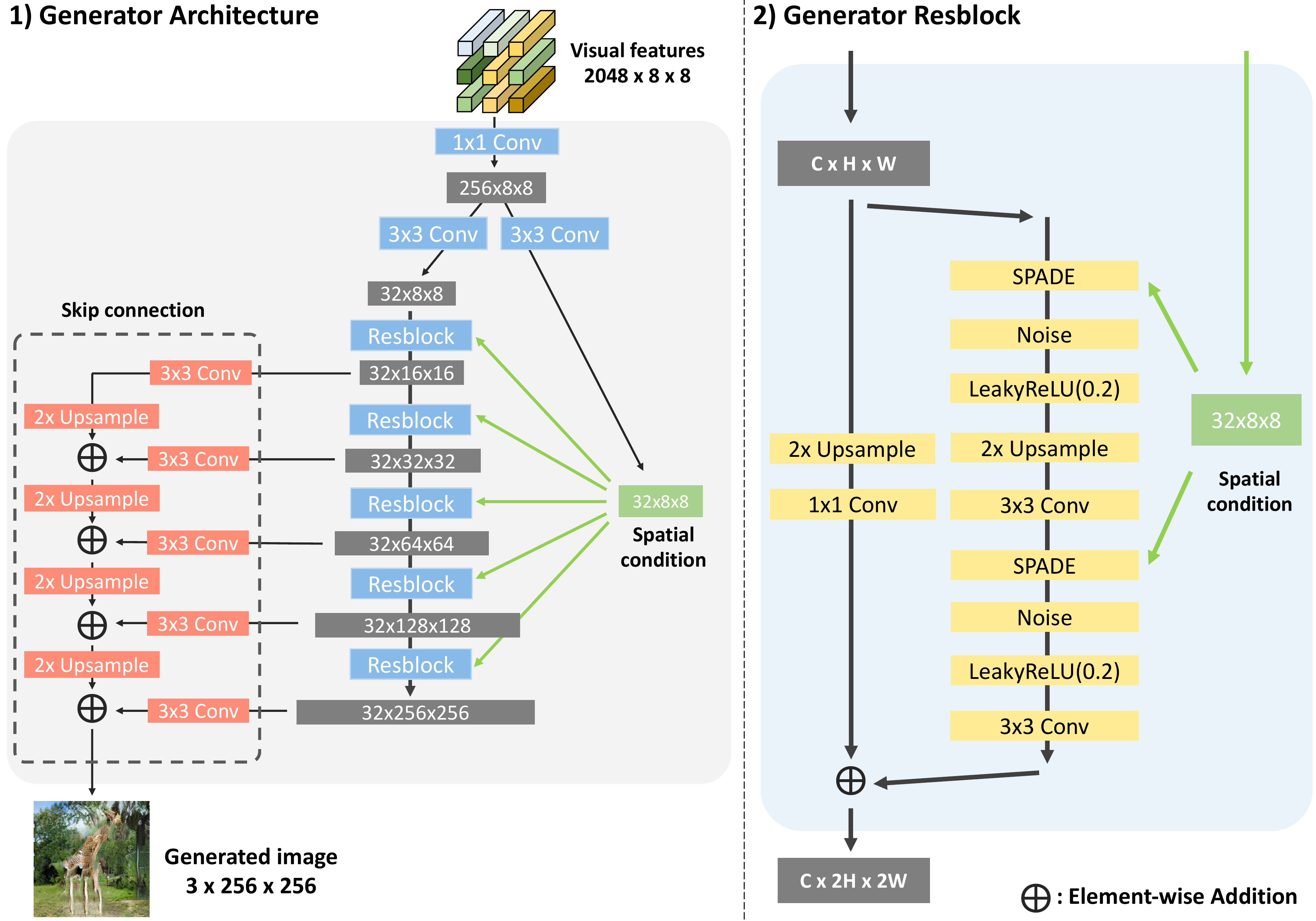}
\end{center}
\caption{
    Generator architecture that takes 8x8 grid visual features and generates 256x256 images.
}
\label{fig:generator_architecture}
\end{figure*}

Our generator consists of multiple residual blocks following SNGAN \cite{Miyato2018a}.
The generator takes (quantized) $8 \times 8$ grid features of Faster R-CNN as input and outputs $256 \times 256$ RGB images.
We use a generator with 5 residual blocks, where each block bilinearly-upsamples feature map by 2.
We use 32 channels of 3x3 kernel for every convolution layer in residual blocks.
Note that many existing generator architectures \cite{Miyato2018a, Wang2018h, Karras2019a, Karras2019} have  residual blocks starting from higher dimensions (eg. 512, 1024) in low-resolution then gradually decrease the dimension as feature maps are spatial upsampled. However, we found that using fixed-sized small dimension for all residual blocks makes training more stable.
Each residual block has spatially adaptive instance norm (SPADE) \cite{Park2019, huang2017b} that guides the residual block using spatial information of $8 \times 8$ grid features.
After each spatially adaptive instance norm, we multiply spatial gaussian noise on feature maps to make model less focus on local texture following StyleGAN \cite{Karras2019a}.
We use spectral normalization \cite{Miyato2018} after each convolution layer in generator.
Following StyleGAN-v2 \cite{Karras2019}, we use skip connection for each residual block to generate final output.
Our generator has 1.7M trainable parameters.
The detailed architecture of our generator is illustrated at Fig. \ref{fig:generator_architecture}.

\begin{figure*}[h!]
\begin{center}
\includegraphics[
                width=\textwidth,
                 ]{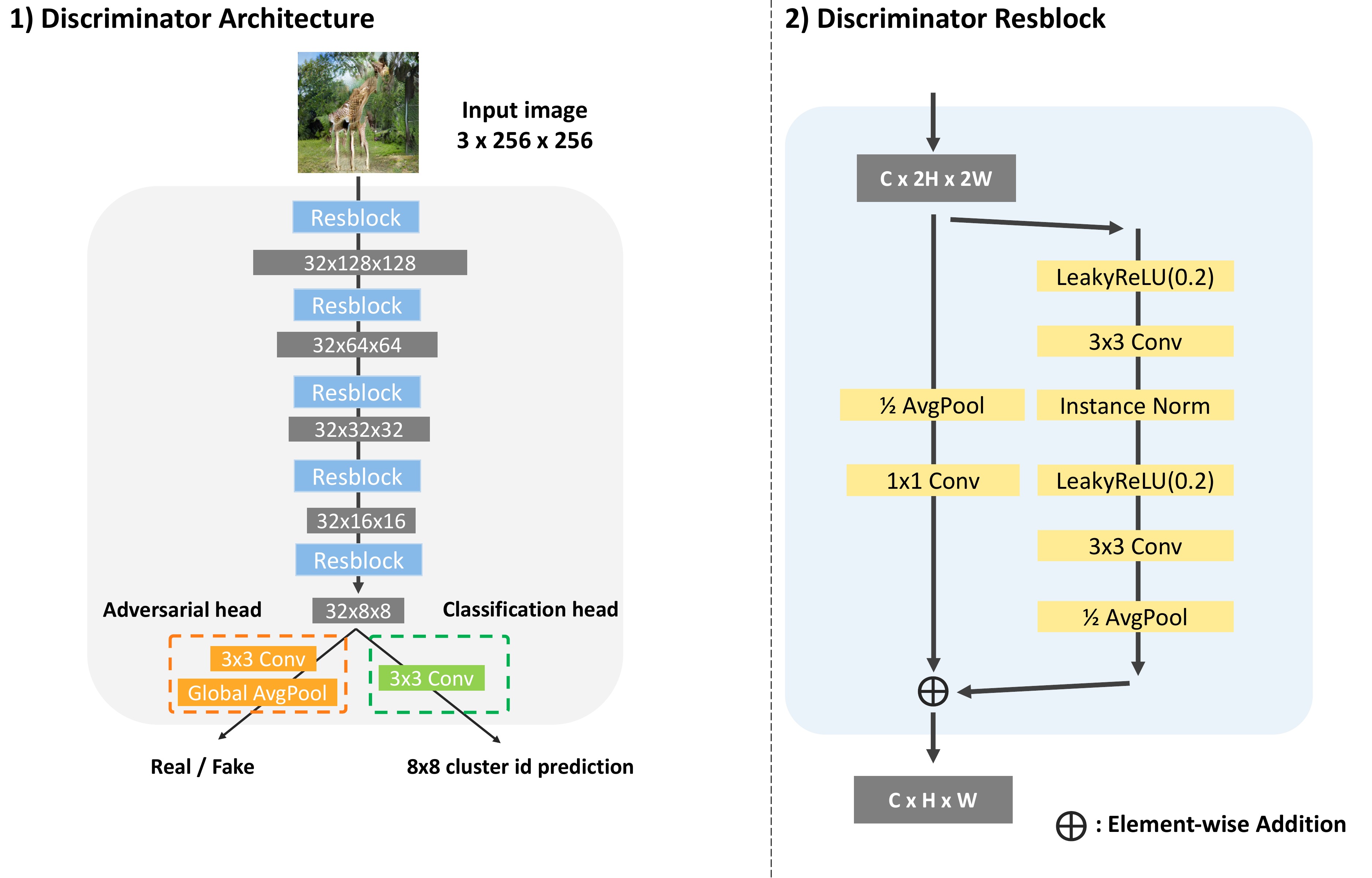}
\end{center}
\caption{
    Discriminator architecture that takes 256x256 images.
}
\label{fig:discriminator_architecture}
\end{figure*}

\subsection{Discriminator Architecture}
Discriminator also consists of multiple residual blocks.
We use a discriminator with 5 residual blocks, where each residual block downsamples feature map by 2.
We use 64 channels of 3x3 kernel for every convolution layer in residual blocks.
We use spectral normalization after each convolution layer in discriminator.
In contrasts to generator, discriminator (1) uses instance norm \cite{Ulyanov2016} instead of adaptive instance norm, (2) does not gaussian noise multiplication and (3) does not use skip connection.
Output of the 5 residual blocks are $8 \times 8$ feature map.
Our discriminator have two heads taking these feature maps:
(1) adversarial head spatially averaging $8 \times 8$ feature map and predicting whether input image is from original image domain or not and (2) classification head predicting cluster ids of $8 \times 8$ spatial layouts from input image. 
Our discriminator has 0.5M trainable parameters.
The detailed architecture of our discriminator is illustrated at Fig. \ref{fig:discriminator_architecture}.

\subsection{Dataset}
We train our model on \coco\ \texttt{train2014} split, which consits of 82783 images.

\subsection{Training}

Our generator and discrminator are trained with 4 losses: (1) hinge adversarial loss \cite{Lim2017, Tran2017}, (2) AC-GAN loss \cite{Odena2017}, (3) discriminator feature match loss \cite{Wang2018h} and (4) perceptual loss \cite{Johnson2016} following \cite{Park2019}.
Following pix2pixHD \cite{Wang2018h}, coefficients for the losses are (1, 1, 10, 10) respectively.
Adversarial loss guides generator to output images close to original images.
The rest of the losses guide generator to output images close to specific target images using spatial layout inputs.
We use ResNet-50 \cite{He2016} for perceptual loss.
Detail of losses are explained in Sec. \ref{sec:generator_loss}.

We use Adam optimizer \cite{Kingma2015} with $(\beta^1, \beta^2)=(0, 0.999)$ and two-time update rule \cite{Heusel2017} with learning rate of 0.0004 and 0.0001 for generator and discriminator respectively.
We train the image generator for 60 epochs with batch size 96.
Training takes 15 hours on 8 NVIDIA Titan V GPUs ($8 \times 12$GB).

\subsection{Losses}
\label{sec:generator_loss}
In below equations, $\hat{X}$ and $X$ refer to generated image and target image respectively.

\paragraph{Adversarial loss}
\begin{align}
    L^{G}_{adv} &= -D^{adv}(\hat{X}) \label{eq:hingeloss_G} \\
    \begin{split}
    L^{D}_{adv} &= max(1 - D^{adv}(\hat{X}), 0) \\
                &+ max(1 - D^{adv}(X), 0) \label{eq:hingeloss_D}    
    \end{split}
\end{align}
where $D^{Adv}$ is discriminator adversarial head.

\paragraph{AC-GAN loss}
\begin{align}
\begin{split}
    L_{ACGAN} &= - \frac{1}{N^2} \sum_{h, w} \log P(D^{cls}_{h,w}(\hat{X})) \\
              &- \frac{1}{N^2} \sum_{h, w} \log P(D^{cls}_{h,w}(X)) \label{eq:ACGANloss}
\end{split}
\end{align}
where $D^{cls}$ is discriminator classification head.

\paragraph{Discriminator feature match loss}
\begin{equation}
    L^{G}_{FM} = \sum_{k} \frac{1}{H^kW^kC^k} \sum_{h,w,c} \ell_{huber} |D^{k}(\hat{X})-D^{k}(X)| \label{eq:Featmatchloss_G} \\
\end{equation}
where
\begin{equation*}
    \ell_{huber}(x) = 
    \begin{cases} 
        0.5 * x^2, \quad if\ |x| \leq 1 \\ 
        |x| - 0.5,   \quad otherwise 
     \end{cases} 
\end{equation*}
and $D^k$ is discriminator's k-th resblock.

\paragraph{Perceptual loss}

\begin{equation}
    L^{G}_{FM-E} = \sum_{k} \frac{1}{H^kW^kC^k} \sum_{h,w,c} \ell_{huber} |E^{k}(\hat{X})-E^{k}(X)| \label{eq:perceptualloss_G} \\
\end{equation}
where $E^k$ is ResNet-50 \cite{He2016}'s k-th resblock 
(\texttt{conv2\_x}, \texttt{conv3\_x}, \texttt{conv4\_x}, \texttt{conv5\_x}).

\paragraph{Total loss}
\begin{align}
    \begin{split}
    L^{G} &= \lambda_{adv} * L^{G}_{adv} \\
            &+ \lambda_{ACGAN} * L_{ACGAN} \\
            &+ \lambda_{FM} * L^{G}_{FM} \\
            &+ \lambda_{FM-E} * L^{G}_{FM-E}
    \end{split} \\
    \begin{split}
    L^{D} &= \lambda_{adv} * L^{D}_{adv} \\
            &+ \lambda_{ACGAN} * L_{ACGAN}
    \end{split}
\end{align}
where $(\lambda_{GAN}, \lambda_{ACGAN}, \lambda_{FM}, \lambda_{FM-E}) = (1,1,10,10)$.

\section{Evaluation details}

\subsection{Image metrics}
To calculate image metrics, we follow \citet{Xu2018e} and randomly sample 30000 images from MS \coco\ \texttt{val2014} split and sample a caption for each image.
Then we generate images from those 30000 captions for each method. We use subset of these 30000 captions for automatic image evaluation.

\begin{table*}[]\footnotesize
  \centering
\resizebox{\textwidth}{!}{
\begin{small}
    \begin{tabular}{l ccccc}
    \toprule
     & R-precision-hard$\uparrow$ & \multicolumn{4}{c}{R-precision-hard categories $\uparrow$} \\
     \cmidrule{3-6}
     & &  Noun & Verb & Color & Number\\
    \midrule
    Original Image                 & 47.6 & 80.4 & 25.3 & 53.4 & 31.4 \\
    \midrule
    DM-GAN \cite{Zhu2019}          & 27.5 & 48.9 & 9.5 & 35.8	& 15.7 \\
    \lxmert                         & 6.6 & 5.6 & 1.7 & 10.2	& 8.7 \\
    \bfseries{\scshape{X-Lxmert}}                        &  25.1 & 41.4 & 9.8 & 30.7 & 18.5\\
    \midrule
    \multicolumn{6}{l}{\bfseries{\scshape{X-Lxmert}}~sampling variations:} \\
    \multicolumn{6}{l}{{\textbf{Autoregressive}}} \\
    \quad TL$\rightarrow$BR                       & 18.9 & 31.6 & 7.3 & 21.0 & 15.5 \\
    \quad Random                               & 24.7 & 41.2 & {10.1} & 28.8 & 18.7 \\
    \quad Random-200                              & 23.3 & 41.2 & {10.1} & 26.5 & 16.5 \\
    \quad Easy-First                              & 22.0 & 35.6 & 8.1 & 25.3 & {18.9} \\
    
    \multicolumn{6}{l}{{\textbf{Parallel}}} \\
    \quad Mask-Predict-1                          & 21.4 & 35.2 & 7.7 & 29.8 & 12.7 \\
    \quad  Mask-Predict-4      & 25.1 & {41.4} & 9.8 & {30.7} & 18.5\\
    \qquad\bfseries{\scshape{= X-Lxmert}} & & & & \\
    \quad Mask-Predict-10                         & 22.6 & 37.3 & 10.0 & 26.1 & 16.9 \\
    \bottomrule
  \end{tabular}
  \end{small}}
  \caption{R-precision-hard per-category scores}
  \label{table:R-prec-hard-cat}
\end{table*}

\paragraph{Inception Score (IS)}
Following \citet{Zhu2019}, we use all 30000 generated images. We use OpenAI implementation\footnote{\url{https://github.com/openai/improved-gan/tree/master/inception_score}} to calculate IS.

\paragraph{Fréchet Inception Distance (FID)}
Following \citet{Zhu2019}, we use all 30000 generated images. We use PyTorch port of official implementation\footnote{\url{https://github.com/mseitzer/pytorch-fid/tree/802da3963113b5b5f8154e0e27580ee4c97460ab}} to calculate FID.

\paragraph{R-precision-easy}
We use all 30000 generated images.
For R-precision-easy, we sample 99 negative captions for each caption, where all negative captions correspond to different \texttt{val2014} images.

\paragraph{R-precision-hard}
For each R-precision-hard category (noun/verb/color/number), we use 1000 randomly sampled caption that contains a category word. Then we generate 9 negative captions by swapping the detected category word with another word with same category.
We use POS-tagging with spaCy\footnote{\url{https://spacy.io/}} to find category words from a caption.
We present per-category score of R-precision-hard at table~\ref{table:R-prec-hard-cat}.

\subsection{Human evaluation}
We use Amazon Mechanical Turk\footnote{\url{https://www.mturk.com/}} for human evaluation.

\begin{figure*}[h]
\begin{center}
\includegraphics[
                width=\textwidth,
                 ]{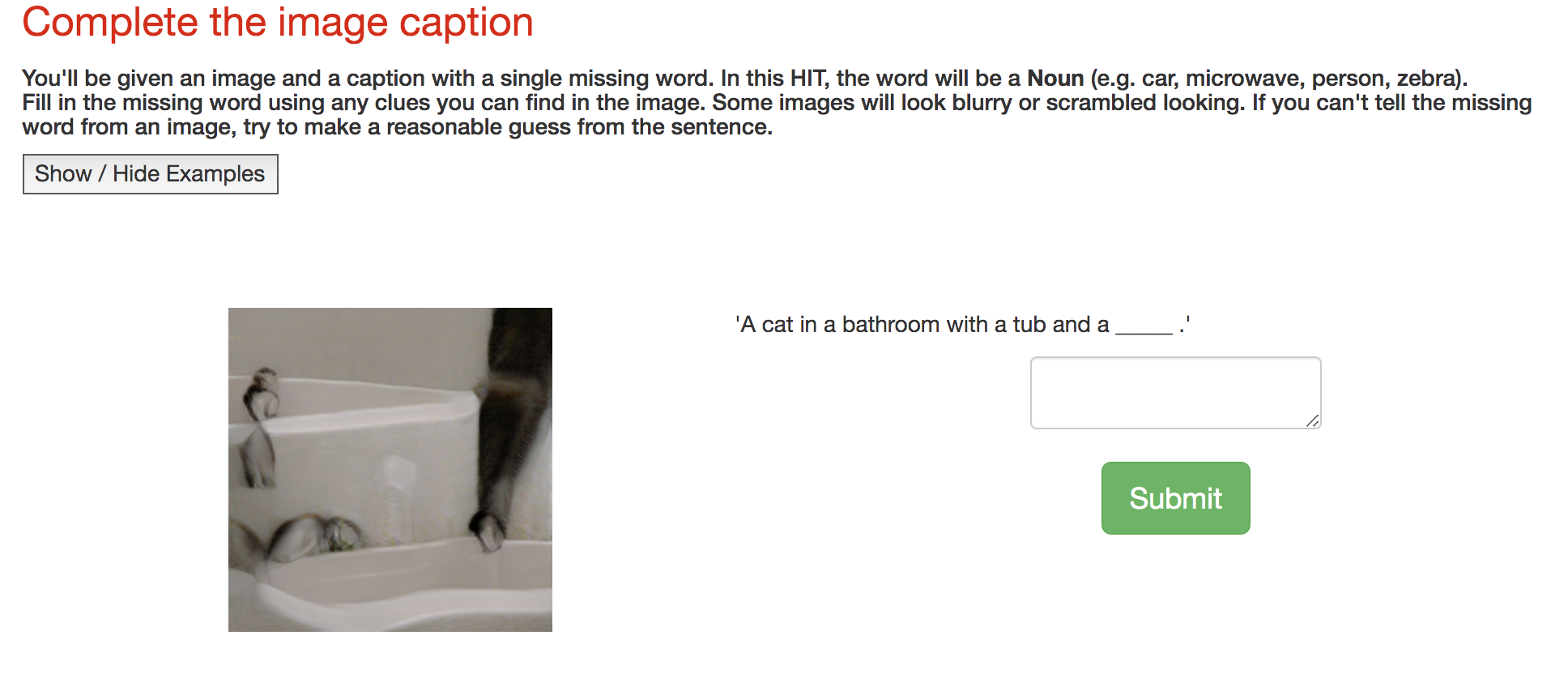}
\end{center}
\vspace{-10pt}
\caption{Screenshot of HUMMUS score evaluation system}
\label{fig:AMT_HUMMUS}
\end{figure*}

\paragraph{HUMMUS score}
For each \hummus\ category (noun/verb/color), we use 100 randomly sampled images. Then we mask out words in the same fashion as in R-precision-hard metric. A total of 280 unique crowdworkers completed the task, with a median of 13 images annotated per worker. 
We present per-category score of \hummus\ score at table~\ref{table:hummus}.
Fig~\ref{fig:AMT_HUMMUS} shows screenshot of \hummus\ score (noun category) evaluation task.

\begin{table}[h!]\footnotesize
  \centering
    \begin{tabular*}{\columnwidth}{lcccc}
    \toprule
    & \hummus$\uparrow$ & \multicolumn{3}{c}{\hummus\ Categories$\uparrow$} \\
    & & Noun & Verb & Color\\
    \midrule
    Original Image       & 0.73 & 0.79 & 0.52 & 0.89\\
    \midrule
    DM-GAN               & \textbf{0.49} & 0.42 & \textbf{0.45} & \textbf{0.60} \\
    \lxmert              & 0.27 & 0.16 & 0.43 & 0.21 \\
    \bfseries{\scshape{X-Lxmert}}             & \textbf{0.49} & \textbf{0.55} & 0.42 & 0.50 \\
    \midrule
    \multicolumn{5}{l}{\bfseries{\scshape{X-Lxmert}} sampling variations:} \\
    TL$\rightarrow$BR    & 0.43 & 0.45 & 0.42 & 0.41 \\
    Random               & 0.48 & 0.52 & 0.42 & 0.51  \\
    Mask-Predict-1       & 0.47 & 0.48 & 0.40 & 0.54 \\
    \bottomrule
  \end{tabular*}
  \caption{Evaluating semantics with \hummus.}
  \vspace{-0.3cm}
  \label{table:hummus}
\end{table}

\paragraph{Pairwise preference}
For \textit{Semantic preference} task, we ask annotators (1) `Which image best matches the caption?' with caption.
For \textit{Fidelity preference} task, we ask annotators `Which image looks more realistic?' without providing the caption.
A total of 357 unique crowdworkers completed the task, with a median of 17 annotations performed per worker. 

\begin{figure*}[h]
\begin{center}
\includegraphics[
                width=\textwidth,
                 ]{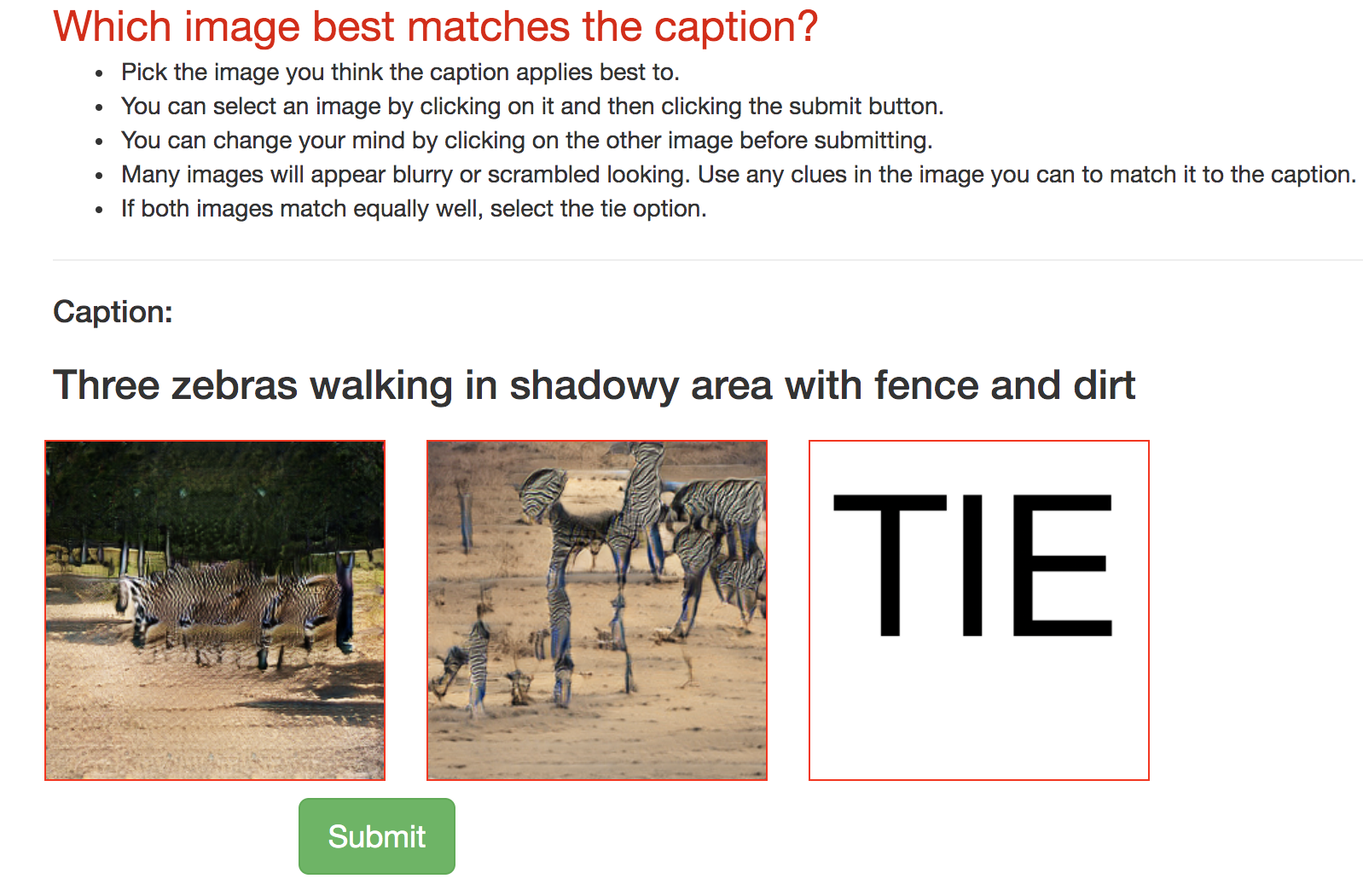}
\end{center}
\vspace{-10pt}
\caption{Screenshot of Semantic preference evaluation system}
\label{fig:AMT_human_with_caption}
\end{figure*}

\begin{figure*}[h]
\begin{center}
\includegraphics[
                width=\textwidth,
                 ]{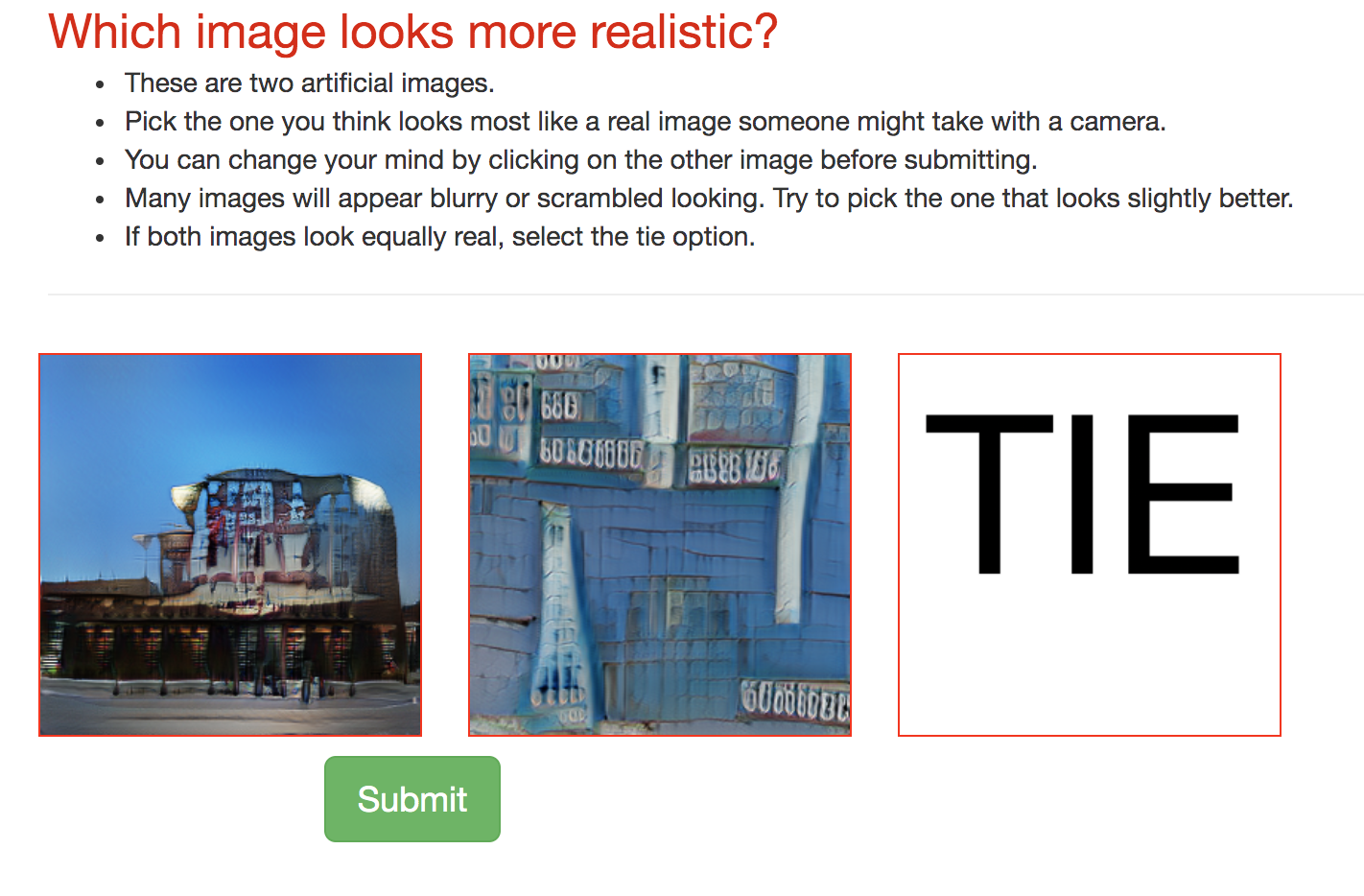}
\end{center}
\vspace{-10pt}
\caption{Screenshot of Fidelity preference evaluation system}
\label{fig:AMT_human_without_caption}
\end{figure*}

Fig~\ref{fig:AMT_human_with_caption} shows screenshot of Semantic preference evaluation task, and
Fig~\ref{fig:AMT_human_without_caption} shows screenshot of Fidelity preference evaluation task.

\end{document}